\documentclass[pdflatex,sn-mathphys-num]{sn-jnl}

\usepackage{graphicx}%
\usepackage{multirow}%
\usepackage{amsmath,amssymb,amsfonts}%
\usepackage{amsthm}%
\usepackage{mathrsfs}%
\usepackage[title]{appendix}%
\usepackage{xcolor}%
\usepackage{textcomp}%
\usepackage{manyfoot}%
\usepackage{booktabs}%
\usepackage{algorithm}%
\usepackage{algorithmicx}%
\usepackage{algpseudocode}%
\usepackage{listings}%
\usepackage{marvosym}
\usepackage{hyperref}

\usepackage{xspace}
\usepackage{adjustbox}
\usepackage{caption}
\usepackage{tabularx}
\usepackage{multirow}
\usepackage{booktabs}

\usepackage{colortbl}
\usepackage{adjustbox}

\input{def.set}

\theoremstyle{thmstyleone}%
\theoremstyle{thmstyletwo}%

\theoremstyle{thmstylethree}%

\newlength\savewidth

\begin{document}

\title[Article Title]{CT-GLIP: 3D Grounded Language-Image Pretraining with CT Scans and Radiology Reports for Full-Body Scenarios}

 
\author[1,2]{\fnm{Jingyang} \sur{Lin}}

\author[1]{\fnm{Yingda} \sur{Xia}}

\author[1,3]{\fnm{Jianpeng} \sur{Zhang}}

\author[1,3]{\fnm{Ke} \sur{Yan}}

\author[4]{\fnm{Kai} \sur{Cao}}

\author[1]{\fnm{Le} \sur{Lu}}

\author*[2]{\fnm{Jiebo} \sur{Luo}}

\author*[1]{\fnm{Ling} \sur{Zhang}}

\affil[1]{\orgname{DAMO Academy, Alibaba Group}}
\affil[2]{\orgname{University of Rochester}}
\affil[3]{\orgname{Hupan Lab}, \postcode{10587}, \state{Hangzhou}, \country{China}}
\affil[4]{\orgname{Department of Radiology, Shanghai Institution of Pancreatic Disease}, \state{Shanghai}, \country{China}}

\makeatletter
\DeclareRobustCommand\onedot{\futurelet\@let@token\@onedot}
\def\@onedot{\ifx\@let@token.\else.\null\fi\xspace}
\makeatother

\makeatletter
\def\eg{\emph{e.g}\onedot} \def\Eg{\emph{E.g}\onedot}
\def\ie{\emph{i.e}\onedot} \def\Ie{\emph{I.e}\onedot}
\def\cf{\emph{c.f}\onedot} \def\Cf{\emph{C.f}\onedot}
\def\etc{\emph{etc}\onedot} \def\vs{\emph{vs}\onedot}
\def\wrt{w.r.t\onedot} \def\dof{d.o.f\onedot}
\def\etal{\emph{et al}\onedot}


\abstract{
3D medical vision-language (VL) pretraining has shown potential in radiology by leveraging large-scale multimodal datasets with CT-report pairs.
However, existing methods primarily rely on a global VL alignment directly adapted from 2D scenarios. The entire 3D image is transformed into one global embedding, resulting in a loss of sparse but critical semantics essential for accurately aligning with the corresponding diagnosis.
To address this limitation, we propose CT-GLIP, a 3D Grounded Language-Image Pretrained model that constructs fine-grained CT-report pairs to enhance \textit{grounded} cross-modal contrastive learning, effectively aligning grounded visual features with precise textual descriptions.
Leveraging the grounded cross-modal alignment, CT-GLIP improves performance across diverse downstream tasks and can even identify organs and abnormalities in a zero-shot manner using natural language.
CT-GLIP is trained on a multimodal CT dataset comprising 44,011 organ-level CT-report pairs from 17,702 patients, covering 104 organs.
Evaluation is conducted on four downstream tasks: zero-shot organ recognition (OR), zero-shot abnormality detection (AD), tumor detection (TD), and tumor segmentation (TS). Empirical results show that it outperforms its counterparts with global VL alignment.
Compared to vanilla CLIP, CT-GLIP achieves average performance improvements of 15.1\% of F1 score, 1.9\% of AUC, and 3.2\% of DSC for zero-shot AD, TD, and TS tasks, respectively.
This study highlights the significance of grounded VL alignment in enabling 3D medical VL foundation models to understand sparse representations within CT scans.
}

\begingroup
\renewcommand\thefootnote{}
\footnotetext{$^*$~Corresponding Authors: 
Jiebo Luo (\href{mailto:jluo@cs.rochester.edu}{jluo@cs.rochester.edu}); 
Ling Zhang (\href{mailto:fabiozhang0722@gmail.com}{fabiozhang0722@gmail.com}).}
\addtocounter{footnote}{-1}
\endgroup

\maketitle

\section{Introduction}

Computed Tomography (CT) analysis plays a crucial role in modern medicine, enabling non-invasive detection of conditions like tumors and fractures without requiring surgery~\cite{costantini2024advancements,shaffie2021novel,aerts2014decoding,lambin2017radiomics,ardila2019end}. CT analysis is also instrumental in treatment planning, offering detailed anatomical information that guides therapies~\cite{wang2019radiomics,louis2024identification,vazquez2024clinical}. Compared to other 3D imaging techniques, such as Magnetic Resonance Imaging (MRI), CT scans are typically more cost-effective due to lower equipment costs and faster scan times~\cite{healthiamges2019mrivsct,mohapatra2023localization,pickhardt2023opportunistic,pickhardt2020automated}.
However, the increasing amount of CT examinations, which exceeds 93 million in 62 million patients annually in the United States~\cite{statnews2023ct,smith2025projected}, has made manual analysis increasingly impractical, especially as the number of radiologists has remained relatively stable~\cite{acr2024how}. 
To alleviate the labor burden, automated systems for CT analysis are urgently needed. Artificial Intelligence (AI) has recently emerged as a powerful tool, enabling automated CT analysis with high accuracy and efficiency~\cite{isensee2021nnu,hosny2018artificial,cao2023large,primakov2022automated}.

AI algorithms have shown their potential in understanding CT imaging~\cite{isensee2021nnu,cao2023large,hu2025ai,mei2020artificial,chen2024transunet,ardila2019end,dosovitskiy2021an}. With the development of deep learning~\cite{krizhevsky2012imagenet,he2016deep,vaswani2017attention,ronneberger2015u} and the emergence of well-annotated large-scale data~\cite{hamamci2024developing,qu2023abdomenatlas,yan2018deeplesion}, AI algorithms have met and even exceeded expert-level performance in specific applications for CT analysis, such as disease detection, diagnosis, and 3D semantic segmentation~\cite{esteva2017dermatologist,de2018clinically,ardila2019end,topol2019high,mckinney2020international,lotter2021robust}.
However, despite these advancements, the effectiveness of supervised learning approaches heavily relies on the availability of large-scale and high-quality annotated data. The high cost of data annotation presents a major challenge, limiting the scalability and further development of these methods for CT analysis.
Therefore, relying solely on supervised learning remains impractical.

To overcome the bottleneck of high annotation costs from supervised learning, weakly supervised learning provides a promising alternative by reducing dependence on extensively labeled data~\cite{lu2021data,radford2021learning,jiang2024transformer,jia2021scaling,zhai2023sigmoid}.
Unlike supervised learning, weakly supervised methods intend to train models with limited or imprecise annotations, which provide partial or noisy supervision without requiring manual labeling.
CLIP (Contrastive Language-Image Pretraining) method~\cite{radford2021learning} is an early attempt at weakly supervised learning for 2D images in general scenarios. It leverages weak labels (\ie, noisy or imprecise vision-language pairs) to facilitate vision-language alignment, enabling models to learn meaningful visual representations without manual annotations.
Building on the success of CLIP, previous works~\cite{zhang2022contrastive,huang2021gloria,wang2022medclip} in the medical imaging domain have adapted it for 2D medical imaging (\eg, X-ray images), leveraging naturally occurring paired medical diagnostic reports to learn meaningful medical visual representations.
Beyond applying weakly supervised learning to 2D medical VL alignment, recent works~\cite{hamamci2024developing, blankemeier2024merlin} have extended it to 3D medical imaging, particularly in volumetric modalities like CT scans.
CT-CLIP~\cite{hamamci2024developing} aims to provide generalist foundation capabilities for 3D medical imaging and reduce reliance on task-specific supervision, employing contrastive learning techniques~\cite{oord2018representation} to achieve supervised-level performance for multi-abnormality detection in a zero-shot manner.Merlin~\cite{blankemeier2024merlin} integrates both structured electronic health records (EHR) and unstructured radiology report information as weak supervision signals. Optimized by contrastive learning, Merlin can effectively capture 3D medical visual representations, enhancing the understanding of volumetric medical data.
With 3D VL alignment, pretrained vision-language models (VLMs) demonstrate remarkable generalization ability across various medical downstream tasks, even without task-specific training~\cite{hamamci2024developing,blankemeier2024merlin}. This capability highlights the potential of weakly supervised learning in enabling more efficient and generalizable AI-driven 3D medical analysis.

Although 3D VL alignment has significantly advanced automated CT analysis, existing works~\cite{hamamci2024developing,blankemeier2024merlin} primarily rely on \emph{global alignment}, treating an entire CT volume and its report as a single contrastive pair. Unlike 2D natural images, CT volumes are inherently information-sparse and anatomically complex. This \emph{global} paradigm collapses hundreds of slices into one representation, resulting in a significant information loss of sparse yet clinically salient regions.
As a result, global alignment often suffers from imprecise supervision and weak correspondences, which can hinder the model from learning accurate and discriminative representations.

To address these limitations, grounded vision-language (VL) alignment is necessary. Unlike global alignment, it pairs anatomical regions with their corresponding textual findings or diagnoses, which satisfies the clinical need for anatomically precise CT analysis. Such a fine-grained matching paradigm eliminates grounded information loss, strengthens supervision signals, and improves detailed understanding.
To this end, we introduce CT-GLIP (Grounded Language-Image Pretraining with CT scans), an innovative approach to grounded VL alignment in CT analysis.
CT-GLIP is trained on fine-grained vision-language pairs at the organ-level, achieving precise alignment between detailed visual components and textual descriptions.
As illustrated in Fig.~\hyperref[fig:overview]{1a}, our approach first preprocesses CT scans using Totalsegmentator~\cite{wasserthal2023totalsegmentator}, segmenting 104 distinct organs, and employs GPT-3.5 Turbo~\cite{openai_gpt35_turbo_0125} followed by manual verification to extract organ-specific descriptions from radiology reports. This procedure produces a multimodal CT-Report dataset for grounded VL alignment with 44,011 fine-grained VL pairs from 17,702 patients covering 104 organs.
To facilitate grounded cross-modal alignment, CT-GLIP utilizes two primary objectives: (1) \emph{anatomy contrastive learning}, enhancing basic medical visual-textual understanding, and (2) \emph{diagnosis contrastive learning}, enabling effective zero-shot abnormality detection.
Contrastive learning typically suffers from limited diversity when using small mini-batch sizes, leading to biased gradients and reduced performance~\cite{chen2020simple,he2020momentum,oord2018representation,wu2018unsupervised,cai2020joint,wang2022low,wang2021improving}.
To increase the diversity and amount of the contrastive pairs, we further introduce the \emph{abnormality dictionary} comprising diverse textual descriptions of abnormalities sourced from clinical practice. This dictionary effectively simulates larger batch sizes by offering diverse contrastive pairs, thus improving the optimization of contrastive learning~\cite{oord2018representation}.
Empirically, the pretrained models are validated on both zero-shot and fine-tuning evaluations (see Fig.~\hyperref[fig:overview]{1b}): (1) the zero-shot evaluation involves organ recognition and abnormality detection tasks using validation and test datasets with 643 and 1,130 patients, respectively, targeting 104 organs and 16 common abnormalities. (2) Fine-tuning evaluation contains tumor detection and segmentation tasks  (see Fig.~\hyperref[fig:overview]{1c}) encompassing 700 non-contrast CT scans of 700 patients, specifically targeting seven of the most prevalent types of cancer, with 100 patients for each type. 
Extensive experiments demonstrate that CT-GLIP outperforms the models with global CT-report alignment~\cite{radford2021learning,jia2021scaling} in 3D medical imaging. It achieves notable zero-shot performance for organ recognition and abnormality detection, and enhances fine-tuning performance on tumor segmentation and detection for both CNN (Convolutional Neural Network~\cite{krizhevsky2012imagenet,simonyan2015very,he2016deep}) and ViT(Vision Transformer~\cite{dosovitskiy2021an})-based models~\cite{isensee2021nnu,xie2022unimiss}.

This study introduces CT-GLIP, which highlights the significance of grounded VL alignment for the 3D medical VL foundation models.
CT-GLIP achieves more accurate alignment between fine-grained visual and textual representations, addressing the limitations of existing global VL alignment methods~\cite{blankemeier2024merlin,hamamci2024developing,radford2021learning,jia2021scaling}.
Therefore, the CT-GLIP framework promotes stronger 3D vision-language foundation models for CT analysis, which pave the way toward broader adoption of AI-driven methods in clinical workflows, ultimately supporting improved diagnostic accuracy and efficiency.
\section{Results}
In this study, we introduce CT-GLIP, a 3D grounded language-image pretraining framework specifically designed for grounded cross-modal alignment between CT scans and radiology reports.
To facilitate the grounded VL alignment, we construct a multimodal CT-report dataset, involving 4,011 fine-grained VL pairs from 17,702 patients covering 104 organs.
We perform a comprehensive evaluation, starting with zero-shot tasks (\ie, organ recognition and abnormality detection) across validation and test datasets comprising 643 and 1,130 patients, respectively, covering 104 distinct organs and 16 common abnormalities.
Moreover, we conduct fine-tuning evaluations on tumor detection and segmentation tasks using 700 non-contrast CT scans from 700 patients, evenly distributed among seven prevalent cancer types (100 patients per type).

\subsection{Grounded Language-Image Pretraining with Grounded CT-Report Pairs}
To enable effective pretraining of 3D vision-language foundational models for CT analysis, we first collect 20,481 CT-report pairs from 19,693 unique patients. The word cloud of the raw reports is presented in Fig.~\hyperref[fig:data_stat]{2d}. The CT scans span 14 primary anatomical regions (see Fig.~\hyperref[fig:data_stat]{2e}). 
To facilitate the grounded CT-Report alignment, we further preprocess both CT scans and radiology reports into fine-grained cross-modal pairs.
On the visual side, we employ TotalSegmentator~\cite{wasserthal2023totalsegmentator} for generating high-quality segmentation masks for 104 distinct organs, covering full-body scenarios.
For textual grounding, we consider two types of organ-level descriptions: anatomical and diagnostic descriptions. For the anatomical descriptions, we design template-based descriptions~\cite{radford2021learning} that briefly describe each of the 104 segmented organs, supplying the model with a basic understanding of visual–textual concepts within CT scans.
To capture richer diagnostic descriptions, we extract organ-level diagnoses by parsing raw radiology reports with GPT-3.5-turbo~\cite{openai_gpt35_turbo_0125}. These generated diagnoses are manually spot-checked to ensure their correctness.
In this way, we ultimately produce high-quality data from 17,702 patients along with 44,011 pairs of grounded CT scans and clinical diagnoses. We then divide this multimodal dataset into training and test sets for experimental studies, with statistical information reported in Fig.~\hyperref[fig:data_stat]{2f-i}.
This grounded CT-report dataset enables further exploration of cross-modal correlations between visual representations and the corresponding textual embeddings, promoting a deeper understanding of medical knowledge within CT scans.

Built on the proposed dataset, we introduce CT-GLIP, a grounded language–image pretraining framework for 3D medical vision–language models that focuses on \emph{grounded} cross-modal alignment at anatomically localized scales. Instead of aligning an entire CT scan with a whole radiology report~\cite{blankemeier2024merlin,hamamci2024developing}, CT-GLIP ties specific anatomical regions to their corresponding textual findings or diagnoses, meeting the clinical need for high-resolution and anatomically precise CT analysis.
In particular, the CT-GLIP framework employs a dual‑path architecture, including a 3D vision encoder and a text encoder. 
During training, CT‑GLIP employs fine-grained contrastive learning~\cite{oord2018representation} on a weakly-annotated dataset with grounded VL samples balanced between positive (matched) and negative (mismatched) pairs, thereby eliminating the need for manual labeling.
In particular, the optimization consists of two types of fine-grained contrastive learning: (1) {\em anatomy contrastive learning} aligns a specific organ region with a template-based anatomical description. (\eg, ``this is a liver in the CT scans'' in Fig.~\hyperref[fig:method]{3a}) and (2) {\em diagnosis contrastive learning} matches the same region with its corresponding clinical diagnosis (\eg, ``fatty liver'' in Fig.~\hyperref[fig:method]{3b}). Please refer to Section~\ref{sec:method} for more details of the proposed CT-GLIP framework and the grounded CT-Report multimodal dataset.

After pretraining with the CT-GLIP framework, the pretrained model can be directly applied to several multimodal applications without any additional training. In particular, it can identify organ categories (see Fig.~\hyperref[fig:zeroshot]{4a}) and detect abnormalities (see Fig.~\hyperref[fig:zeroshot]{4c}) in a zero‑shot setting from natural‑language prompts.
Moreover, the pretrained model can adapt quickly to downstream tasks, including tumor detection and segmentation, with a small number of manually annotated samples. Detailed experimental results are presented in the subsequent sections.

\subsection{Superiority of CT-GLIP in Zero-shot Evaluation} \label{sec:zero-shot}

In this section, we evaluate the capabilities of CT-GLIP in recognizing organs and detecting abnormalities in a zero‑shot setting using natural‑language prompts without task-specific fine-tuning.

\noindent \textbf{Zero-shot Organ Recognition}.
To assess zero-shot organ recognition, we evaluate whether grounded visual embeddings of CT scans can be reliably matched to the corresponding organ categories using template-based textual prompts. This ability serves as the basic knowledge of associating grounded visual-textual concepts in CT scans.
As illustrated in Fig.~\hyperref[fig:zeroshot]{4a}, a 3D encoder (nnUNet or MiT) followed by organ-level pooling produces 104 organ-specific visual embeddings. Each embedding is compared against 104 text embeddings generated from a simple template, ``\texttt{This is the \{organ\_name\} in the CT scans.}''
The organ category prediction is the one with the highest similarity, and performance is measured by top-1 accuracy.
Vanilla CLIP~\cite{radford2021learning} serves as a baseline to demonstrate the superiority of CT-GLIP.
We reimplement the vanilla CLIP using our proposed CT-report multimodal dataset for a fair comparison.
The results in the Fig.~\hyperref[fig:zeroshot]{4b} show that the CT-GLIP achieves compelling performance on the zero-shot organ recognition task, whereas the global VL alignment baseline (Vanilla CLIP) struggles with this grounded recognition task. In particular, CT-GLIP models achieve 86.9\% top-1 accuracy with nnUNet and 85.4\% with MiT. These results indicate that explicit grounded cross-modal alignment is critical to learn discriminative representations for organ recognition.

\noindent \textbf{Zero-shot Abnormality Detection}. We further evaluate the capability of CT-GLIP on a zero-shot abnormality detection task. Beyond organ recognition, this task requires matching finding-specific prompts to features pooled strictly within relevant anatomical regions, but also distinguishing whether the region is normal or exhibits a specified abnormality from natural-language prompts.
Fig.~\hyperref[fig:zeroshot]{4c} illustrates the inference process of zero-shot abnormality detection, operating as a binary classification task. For each testing CT image, we provide a pair of normal and abnormal text descriptions with corresponding organ masks.
We assess the similarity between the organ-level grounded visual features and both normal and abnormal textual embeddings for each targeted abnormality. The prediction is made based on the higher similarity score.
We report standard metrics for the binary classification task, comprising F1 score, Positive Predictive Value (PPV), Sensitivity, and AUC.
Results in Fig.~\hyperref[fig:zeroshot]{4d-g} show that global alignment on 3D CT–report pairs (vanilla CLIP baseline) learns limited abnormality cues, whereas CT-GLIP effectively captures sparse but informative semantics from 3D representations by leveraging grounded VL alignment.
With nnUNet backbones, CT-GLIP improves F1 by 15.0\% and AUC by 16.4\% over vanilla CLIP; with MiT backbones, the gains are larger, reaching 15.6\% in F1 and 19.5\% in AUC.
To visualize the per-organ performance, we present the AUC scores over different organs in Fig.~\hyperref[fig:zeroshot]{4h}.
The results demonstrate that CT-GLIP using grounded alignment consistently outperforms vanilla CLIP with global alignment on every organ.
Vanilla CLIP lies largely in the 35–55\% range, which demonstrates that grounded signals are easily diluted by global context. In contrast, most CT-GLIP scores concentrate near the 60–80\% ring, with the kidney, pancreas, and lung achieving the highest scores.
The largest gaps appear for the kidney and pancreas, highlighting that the grounded alignment strengthens fine-grained vision–language associations. Moreover, CT-GLIP shows lower variance across organs, indicating more stable generalization.

Please refer to Section~\ref{sec:method} for more implementation details of zero-shot evaluation, along with data statistics of the test samples. Also, Table~\ref{tab:zeroshot} in Appendix~\ref{sec:appendix} presents raw results shown in Fig.~\hyperref[fig:zeroshot]{4}.

\subsection{Ablation Studies on Components of CT-GLIP}
The CT-GLIP framework comprises three key components: anatomy contrastive learning, diagnosis contrastive learning, and the abnormality dictionary. To clarify their individual contributions, we conduct ablation studies with two variants of CT-GLIP:
\begin{itemize}
    \item CT-GLIP$^\dagger$ retains only the diagnosis contrastive learning, removing the anatomy contrastive learning and the abnormality dictionary.
    \item CT-GLIP$^\ddagger$ preserves both anatomy and diagnosis contrastive learning while excluding the abnormality dictionary.
\end{itemize}
The ablation studies are evaluated on the zero-shot abnormality detection task.

\noindent \textbf{Effectiveness of diagnosis contrastive learning}. Comparing vanilla CLIP to CLIP$^\dagger$ (in Fig.~\hyperref[fig:zeroshot]{4d-g}), we observe substantial gains in zero-shot abnormality detection for both CNN- and ViT-based backbones. The improvements show that diagnosis contrastive learning facilitates effective grounded VL alignment over the sparse 3D representations. More specifically, the largest gains appear in Sensitivity (35.48\% for nnUNet and 33.81\% for MiT), indicating that diagnosis contrastive learning markedly reduces false negatives.

\noindent \textbf{Effectiveness of anatomy contrastive learning}. As shown in Fig.~\hyperref[fig:zeroshot]{4b}, incorporating the anatomy contrastive loss enables zero-shot organ recognition.
Furthermore, we examine its impact on zero-shot abnormality detection by comparing CT-GLIP$^\dagger$ (without anatomy contrastive learning) to CT-GLIP$^\ddagger$ (with anatomy contrastive learning). The anatomy contrastive learning consistently improves performance for both CNN- and ViT-based backbones (in Fig.~\hyperref[fig:zeroshot]{4d, 4g}): with nnUNet, CT-GLIP$^\ddagger$ improves F1 by 1.58\% and AUC by 0.76\% over CT-GLIP$^\dagger$; with MiT, F1 increases by 1.23\% and AUC by 0.85\%. These gains indicate that strengthening basic grounded visual–textual concepts helps the model focus on the relevant anatomical regions, making localized abnormalities easier to detect.

\noindent \textbf{Effectiveness of abnormality dictionary}.
The abnormality dictionary aims to increase the diversity and the amount of contrastive pairs, which benefits the contrastive learning~\cite{oord2018representation}. The scale of the abnormality dictionary is set to 512 since a larger scale will no longer benefit the performance. Comparison between CT-GLIP$^\dagger$ and CT-GLIP (in Fig.~\hyperref[fig:zeroshot]{4d-g}) shows that the abnormality dictionary further enhances the performance on zero-shot abnormality detection: with nnUNet, F1 improves by 0.42\% and AUC by 1.87\%; with MiT, F1 improves by 1.36\% and AUC by 1.78\%.

\subsection{Adapting CT-GLIP to Tumor Detection and Segmentation Tasks via Fine-tuning}
To validate the adaptability and performance of our pretrained model in the segmentation and detection of several prevalent types of cancer on noncontrast CT scans, we conduct fine-tuning evaluation in an in-house dataset encompassing 700 non-contrast CT scans of 700 patients, specifically targeting seven of the most prevalent types of cancer, including lung, breast, liver, esophagus, stomach, colorectum, and pancreas cancer.
In this setting, we compare the proposed CT-GLIP with the other two important baseline models, including Scratch (randomly initialized) and Vanilla CLIP.
For a fair comparison, these three models are fine-tuned on the same training split.
The performance is evaluated in terms of the Dice-Sørensen
Coefficient (DSC) for tumor segmentation. As for the patient-level detection of each type of tumor (the presence or absence of each tumor), we use the 3D volume of the respective tumor as the score for the computation of AUC~\cite{zhu2019multi}.

For both CNN- and ViT-based backbones, our CT-GLIP outperforms the baseline models trained from scratch and those fine-tuned with vanilla CLIP training, as shown in Fig.~\ref{fig:ft}.
As shown in Fig.~\hyperref[fig:ft]{5c}, CT-GLIP outperforms the model trained from scratch and fine-tuned from vanilla CLIP by 4.8\% and 1.3\% in mean tumor segmentation dice score, and 7.4\% and 2.2\% in tumor detection AUC score for the nnUNet backbone, respectively. For the MiT backbones, Fig.~\hyperref[fig:ft]{5f} demonstrates that the respective improvements are 13.1\% and 5.1\% for tumor segmentation, and 7.7\% and 1.6\% for tumor detection.
More specifically, per-organ performances are presented in Fig.~\hyperref[fig:ft]{5a-b} for tumor detection and Fig.~\hyperref[fig:ft]{5d-e} for tumor segmentation.
Generally, either pretrained with vanilla CLIP or CT-GLIP can improve the performance by a large margin, illustrating the importance of pre-learned representation for this clinically significant task.
Our superiority over vanilla CLIP further illustrates the efficacy of our method in leveraging visual-textual associations for enhanced tumor-related image representations.

Please refer to Table~\ref{tab:finetune} in Appendix~\ref{sec:appendix} for raw results shown in Fig.~\hyperref[fig:ft]{5}.
\section{Discussions}

In this study, we introduce CT-GLIP (Fig.~\hyperref[fig:method]{3}), a 3D grounded language-image pretraining framework with grounded VL alignment between CT scans and radiology reports. Leveraging a multimodal dataset (Fig.~\hyperref[fig:data_stat]{2}) of 44,011 grounded CT-report pairs from 17,702 patients, we demonstrate that CT-GLIP surpasses several important baseline models across zero-shot and fine-tuning evaluation. In particular, zero-shot evaluation (Fig.~\hyperref[fig:zeroshot]{4}) consists of 104-way organ recognition and abnormality detection for 16 representative abnormalities, conducted on the test set comprising 1,330 patients. Fine-tuning evaluation (Fig.~\hyperref[fig:ft]{5}) includes tumor detection and segmentation tasks over the 7 most prevalent types of cancer.

Automatic CT analysis is an ideal application for weakly supervised learning~\cite{radford2021learning,jia2021scaling}, since hospital information systems routinely archive CT scans and the corresponding formal radiology report as paired records for diagnostic, operational, and legal purposes~\cite{acr2020practice}.
Based on the large-scale CT-report pairs, previous works~\cite{blankemeier2024merlin,hamamci2024developing} have successfully adapted cross-modal alignment techniques~\cite{radford2021learning} to the 3D medical imaging area (\eg, CT scans). The performance of these 3D vision-language (VL) foundation models even outperforms the supervised methods across several downstream tasks~\cite{hamamci2024developing}.
However, these prior approaches directly apply \emph{global} VL alignment to entire CT volumes and full reports, overlooking inherent differences in information density and structure between 2D and 3D images. 
In 3D medical imaging, clinically relevant voxels are scattered across hundreds of slices. Meanwhile, the content of radiology reports is often complex and composite, which depicts multiple findings distributed across different anatomical regions. As a result, \emph{global} alignment dilutes the grounded correspondences between CT scans and reports, which hinders the learning of discriminative representations.
To address this limitation, we introduce \emph{grounded} cross-modal alignment for CT-report pairs, producing the CT-GLIP framework.
Unlike the global counterparts, CT-GLIP focuses on associating anatomical regions with their corresponding textual findings, which is essential for capturing locally discriminative representations.
Empirically, we conduct extensive experiments on both zero-shot and fine-tuning evaluation settings to demonstrate the superiority of grounded VL alignment.
In zero-shot abnormality detection, CT-GLIP significantly reduces missed abnormal organs compared with the global counterpart (Fig.~\hyperref[fig:zeroshot]{4f}), indicating it effectively captures sparse grounded cues.
When adapted to tumor segmentation and detection, CT-GLIP outperforms both Scratch and vanilla CLIP baselines, highlighting that grounded alignment provides stronger generalization ability.
Beyond performance superiority, ablation studies (Fig.~\hyperref[fig:zeroshot]{4d-g}) on three components of CT-GLIP further provide several insightful findings.
The anatomy contrastive learning establishes reliable anchors on specific anatomical regions, enabling the model to more easily detect abnormalities.
The abnormality dictionary enriches clinical language coverage by increasing the number of contrastive pairs, enabling the model to handle real-world clinical terminology.

Despite the above advancements, there are several limitations regarding our dataset and model.
First, although our dataset of 44,011 grounded CT–report pairs is relatively large by 3D standards, it remains orders of magnitude smaller than 2D vision–language datasets, such as the 400 million image–text pairs for CLIP~\cite{radford2021learning}.
At the current data scale, we cannot reliably characterize scaling behavior for grounded contrastive learning in 3D scenarios.
Future research could extend our approach by expanding the grounded CT–report samples to figure out the scaling behavior of grounded 3D vision–language learning.
Second, Fig.~\hyperref[fig:data_stat]{2f} shows that the proposed dataset consists of diverse anatomical regions across the full body. However, the training distribution is dominated by a few regions ($\sim$60\% lung, $\sim$10\% liver, and $\sim$6\% kidney), which may cause biases in representation learning and downstream performance.
To overcome this data limitation, future work could investigate the extent to which the anatomical imbalance issue affects pretraining and downstream tasks.
Third, to facilitate grounded cross-modal alignment, we employ TotalSegmentator~\cite{wasserthal2023totalsegmentator} to generate pseudo-masks and extract organ-level visual representation. As a result, the segmentation errors produced by TotalSegmentator cannot be corrected during pretraining since this mask generation strategy is fixed and independent of the training framework.
Future research could extend our approach by replacing fixed pseudo-masks with learnable masks that are jointly optimized with the entire vision-language model in an end-to-end training manner.
Fourth, our study only focuses on CT scans. Extending grounded 3D vision–language pretraining to MRI (Magnetic Resonance Imaging), PET-CT (Positron Emission Tomography–Computed Tomography), or 3D ultrasound images is a promising direction. It requires handling higher resolution and longer volumetric context, which will introduce substantial memory costs and make long-range dependency modeling in 3D more challenging. We encourage future work to explore more efficient 3D medical imaging encoders to better handle high-resolution and long-context inputs. Moreover, developing unified multimodal 3D encoders for MRI, PET-CT, and 3D ultrasound might further enhance generalization across diverse 3D imaging modalities.

In conclusion, this study highlights the significance of grounded cross-modal alignment for 3D medical vision-language foundation modals.
Our proposed CT-GLIP overcomes sparse information challenges in 3D medical imaging and shows promise in zero-shot organs recognition and abnormality detection, with implications for improving the downstream task of multi-cancer screening.
This work moves a step closer toward practical, automated CT analysis and lays the groundwork for future research on grounded 3D medical vision-language learning.

\begin{figure}
    \centering
    \includegraphics[width=1\textwidth]{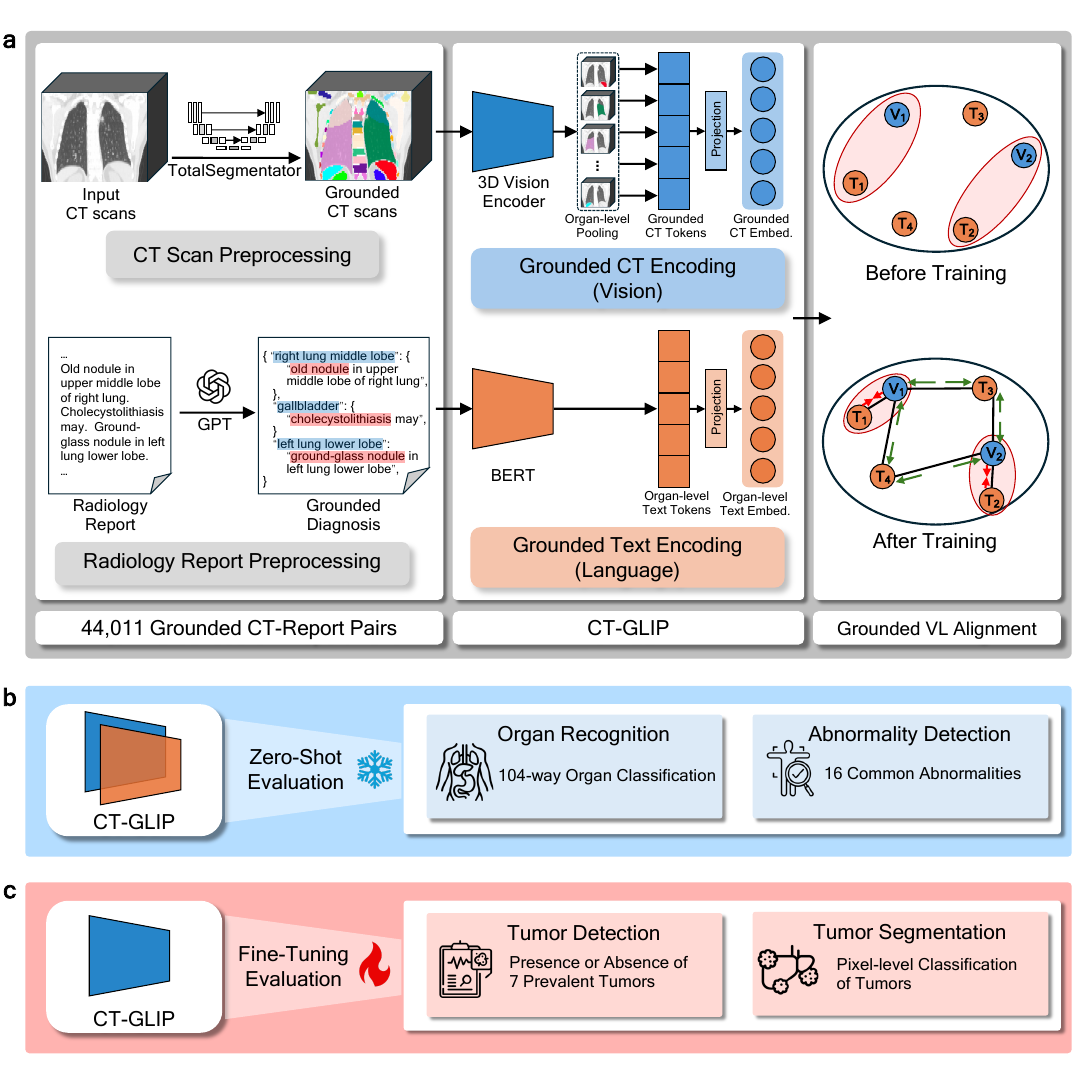}
    \caption{\textbf{Overview of our grounded CT-report multimodal dataset, the CT-GLIP framework, and evaluation protocols}. \textbf{a.} The CT-GLIP framework is trained on 44,011 grounded CT-report pairs from 17,792 patients, covering 104 organs. This fine-grained multimodal dataset enables precise and effective contrastive learning for analyzing CT scans. \textbf{b.} Zero-shot evaluation includes 104-way organ recognition and abnormality detection, conducted on test sets comprising 643 and 1,330 patients, respectively, and targeting 16 common abnormalities across 7 organs. \textbf{c.} Fine-tuning evaluation focuses on tumor detection and segmentation tasks.}\label{fig:overview}
\end{figure}
\begin{figure}
    \centering
    \includegraphics[width=1\textwidth]{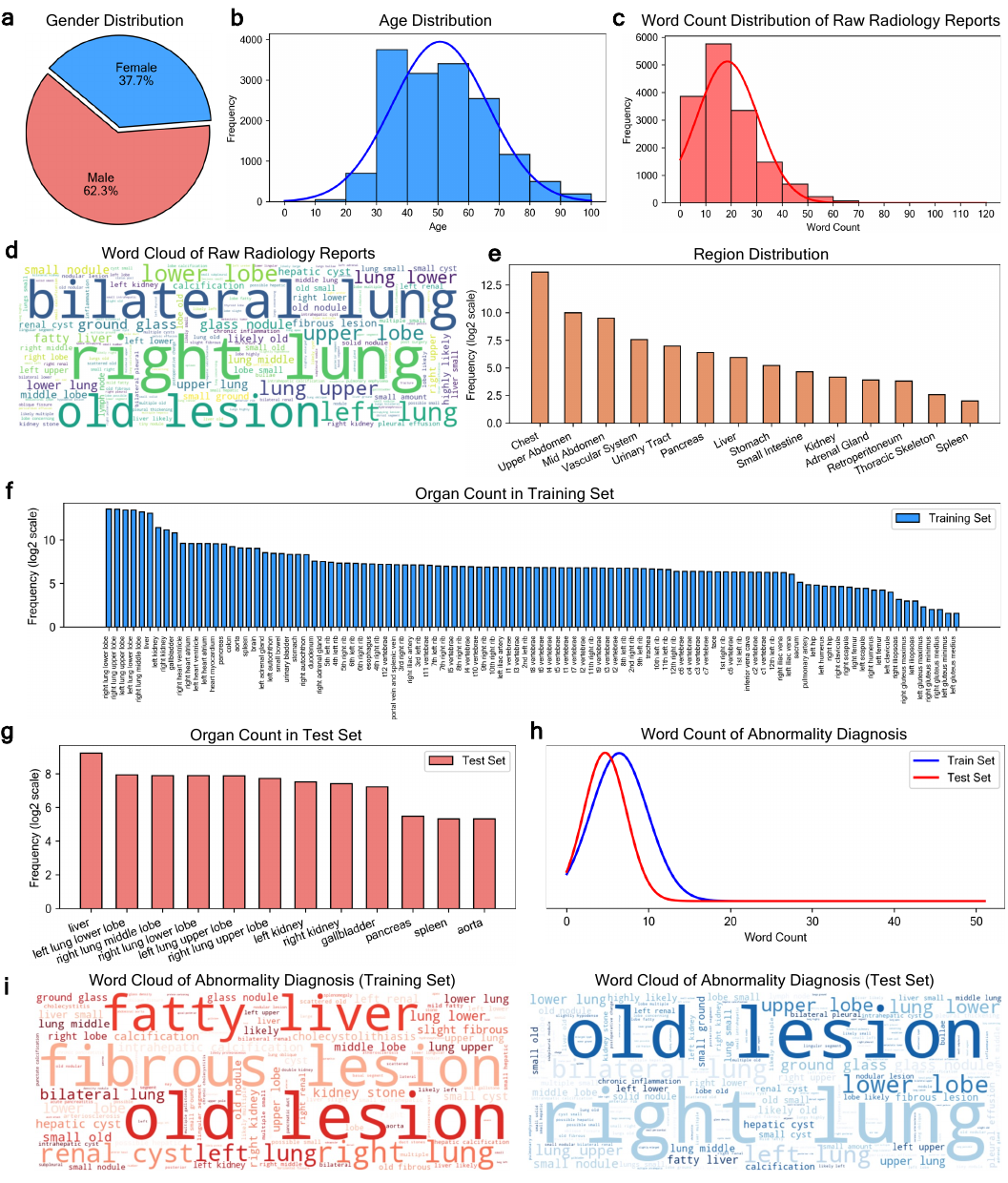}
    \caption{\textbf{Comprehensive statistics of our grounded CT-report multimodal dataset}. \textbf{a.} Gender distribution, with 62.3\% male and 37.7\% female patients. \textbf{b.} Age distribution depicted as a histogram with ten-year bins and an overlaid Gaussian kernel–density estimate, indicating a broad adult cohort with an average age of approximately 50 years. \textbf{c.} Word count distribution of the raw radiology reports. \textbf{d.} Anatomic region frequencies in log scale. \textbf{e.} Word cloud of the raw reports, where font size reflects term frequency. \textbf{f.} Organ frequencies (log scale) in the training set. \textbf{g.} Organ frequencies (log scale) in the test set. \textbf{h.} Word count of the structured abnormality diagnoses for training (blue) and test (red) sets presents a similar distribution. \textbf{i.} Word clouds of the abnormality diagnoses for the training (left) and test (right) sets.}\label{fig:data_stat}
\end{figure}
\begin{figure}
    \centering
    \includegraphics[width=1\textwidth]{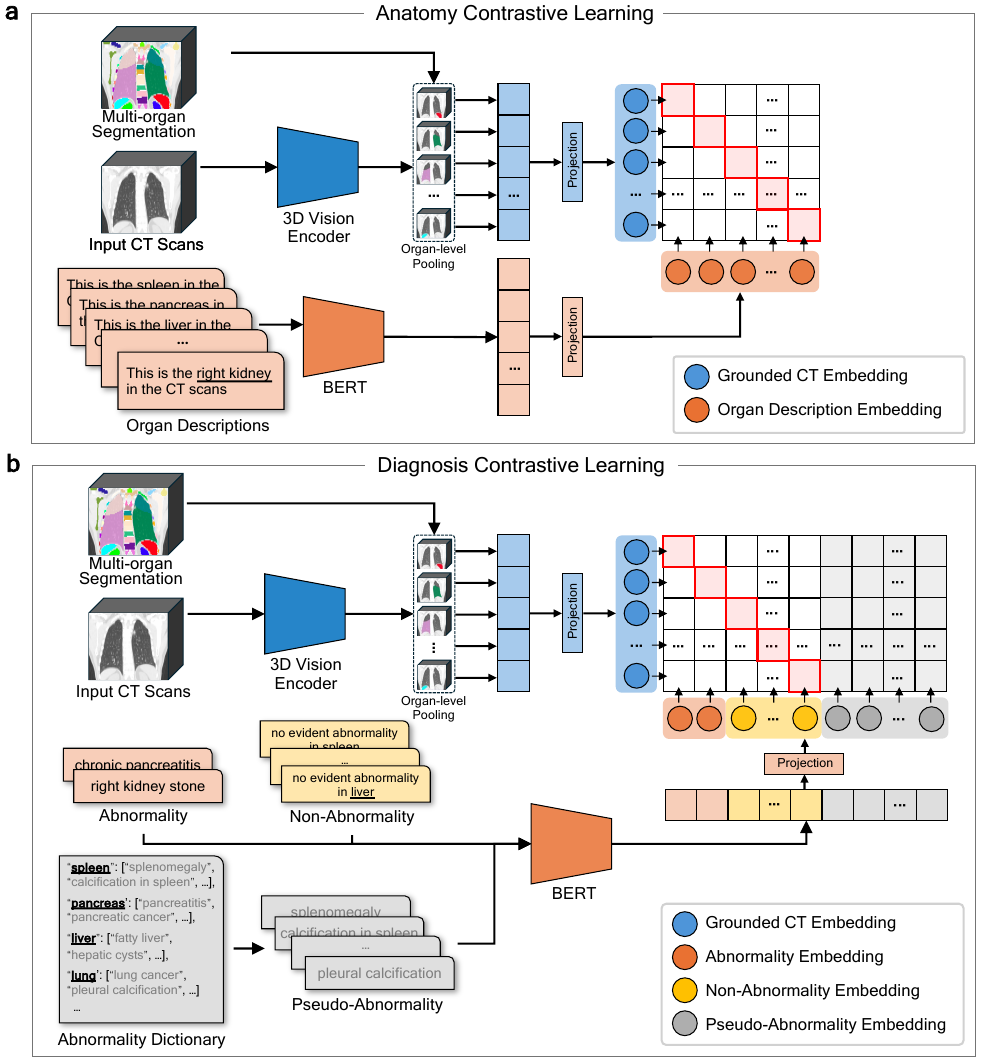}
    \caption{\textbf{Overview of the CT-GLIP framework}, consisting of anatomy and diagnosis contrastive learning. \textbf{a.} \textbf{Anatomy contrastive learning} obtains grounded CT embedding via organ-level pooling with pseudo masks. Then, it pairs each organ with a template description encoded by an expert text encoder to learn anatomical VL alignment. \textbf{b.} \textbf{Diagnosis contrastive learning} builds diagnosis descriptions for each organ (real findings for abnormal, templated ``no evident abnormality'' for normal). Moreover, an additional \textbf{abnormality dictionary} increases the diversity of abnormality descriptions, expanding contrastive pairs and improving  discrimination between abnormality and non-abnormality.}\label{fig:method}
\end{figure}
\begin{figure}
    \centering
    \includegraphics[width=0.925\textwidth]{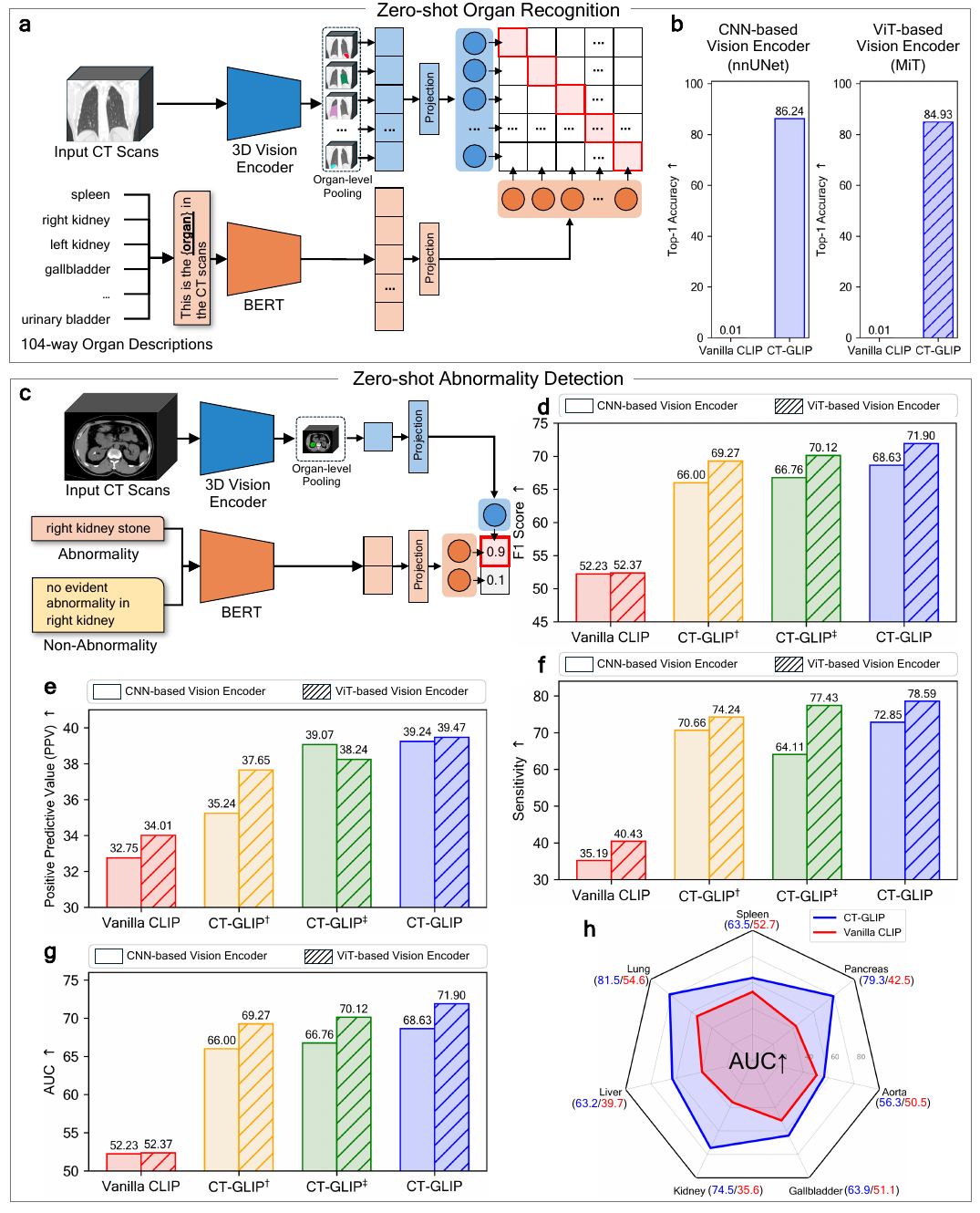}
    \caption{\small \textbf{Zero-shot Evaluation}.
    \textbf{a.} At zero-shot organ recognition inference, 104 organ descriptions are templated and encoded by an expert text encoder. Meanwhile, organ-level visual embeddings are extracted from grounded CT scans by a 3D encoder. Then, the nearest text–image embedding towards a given grounded visual embedding determines the prediction, enabling 104-way zero-shot classification.
    \textbf{b.} Top-1 accuracy with CNN and ViT backbones shows that CT-GLIP achieves strong performance on zero-shot organ recognition, whereas the global VL-alignment baseline (the Vanilla CLIP) struggles on this task.
    \textbf{c.} Zero-shot abnormality detection inference pipeline: organ-level features are contrasted against abnormality and non-abnormality prompts to determine the abnormality label.
    \textbf{d-g.} Zero-shot abnormality detection across CNN- and ViT-based encoders evaluated by F1, Positive Predictive Value, Sensitivity, and AUC: CT-GLIP and its variants ($\dagger$ and $\ddagger$) improve over the Vanilla CLIP, with the \emph{full} CT-GLIP achieving the strongest results.
    \textbf{h.} Per-organ AUC radar plot highlights consistent improvements of CT-GLIP over the Vanilla CLIP across all seven organs.}\label{fig:zeroshot}
\end{figure}
\begin{figure}
    \centering
    \includegraphics[width=1\textwidth]{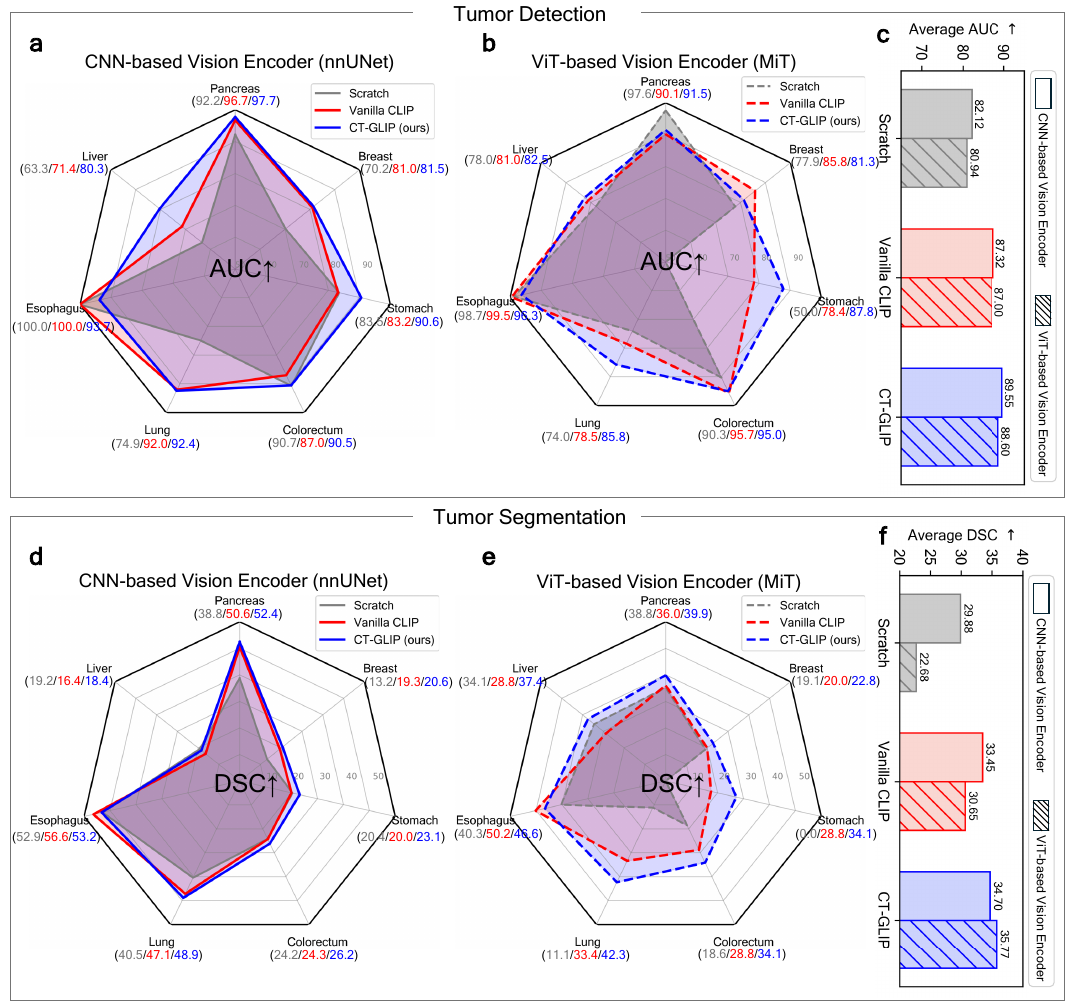}
    \caption{\small \textbf{Fine-tuning Evaluation}.
    \textbf{a.} Per-organ tumor detection is evaluated with a CNN backbone (nnUNet) using AUC. Both pretraining strategies surpass training from scratch, and CT-GLIP consistently outperforms the Vanilla CLIP across most organs.
    \textbf{b.} Using a ViT backbone (MiT),  we obtain the same pattern, highlighting backbone-agnostic benefits from pretraining and the stronger encoder learned by CT-GLIP.
    \textbf{c.} Overall performance on tumor detection of both CNN and ViT backbones confirms that pretraining provides clear improvements, and grounded VL alignment outperforms global VL alignment.
    \textbf{d.} Per-organ tumor segmentation with nnUNet, evaluated by Dice-Sørensen Coefficient (DSC), shows improvements from pretraining, and CT-GLIP provides additional gains beyond the Vanilla CLIP.
    \textbf{e.} Per-organ performance on tumor segmentation with MiT shows that CT-GLIP gains broad improvements on most organs over both Scratch and the vanilla CLIP baselines.
    \textbf{f.} Overall performance on tumor segmentation using both CNN and ViT backbones demonstrates the superiority of CT-GLIP.}\label{fig:ft}
\end{figure}

\clearpage

\section{Methods}\label{sec:method}
\subsection{Data Preparation of Grounded CT-Report Pairs}

\textbf{Data Collection}.
To construct the grounded CT-report dataset, we first collect 20,481 CT-report pairs from 19,693 patients in a single hospital database. The dataset includes 62.3\% male and 37.7\% female patients (Fig.~\hyperref[fig:data_stat]{2a}), where the ages range from 3 to 100 years with a mean of 50.68 and a median of 50 (Fig.~\hyperref[fig:data_stat]{2b}).
For raw radiology reports, we use the impression section of each report since it concentrates the key findings and diagnostic significance. The word count distribution of reports (Fig.~\hyperref[fig:data_stat]{2c}) shows an average of 18.4 words. The word cloud (Fig.~\hyperref[fig:data_stat]{2d}) illustrates broad coverage across lungs, liver, kidneys, spleen, and bones, along with common pathologies such as nodules, cysts, lesions, effusions, and inflammations. It also reflects clinical contexts like postoperative changes and chronic conditions.
The CT scans cover 14 primary anatomical regions (Fig.~\hyperref[fig:data_stat]{2e}), where chest findings form the largest portion, especially in the lungs and thorax. The abdomen is the second largest portion. Smaller subsets from the vascular, urinary, endocrine, and skeletal systems improve the data diversity.

\noindent \textbf{Data Preprocessing for CT Scans and Radiology Report}.
Radiologists typically focus on specific regions to judge whether an abnormality is present. To mirror this clinical practice, we generate segmentation masks for 104 organs using TotalSegmentator~\cite{wasserthal2023totalsegmentator} (Fig.~\hyperref[fig:overview]{1a}, left). These masks define regions of interest for organ-level visual feature extraction, directing the model to the relevant anatomy while filtering out background and other noise.
To obtain grounded textual guidance, we prompt the GPT-3.5-turbo~\cite{openai_gpt35_turbo_0125} model to extract organ–level diagnoses. These grounded diagnoses are then manually spot-checked to ensure their correctness. In addition, we translate the original Chinese radiology report into English to satisfy the language requirement of the ClinicalBERT~\cite{alsentzer2019publicly} model.
After that, we match each organ–level diagnosis with its corresponding organ mask, obtaining grounded CT–report pairs.

\noindent \textbf{Grounded CT-Report Dataset}.
To enable pretraining of grounded cross-modal alignment, we construct a grounded CT-report dataset, which consists of 44,011 organ-level CT-report pairs from 17,702 patients, covering 104 organs. Fig.~\hyperref[fig:data_stat]{2f} presents the organ distribution for the training set.
To evaluate the zero-shot capabilities of CT-GLIP, we further construct validation and test datasets with 643 and 1,130 patients, respectively, targeting 16 common abnormalities across 7 organs. All validation and test samples are manually reviewed, producing a more balanced organ distribution (Fig.~\hyperref[fig:data_stat]{2g}).
We also visualize the number of abnormality diagnoses (Fig.~\hyperref[fig:data_stat]{2h}) for both training and test sets, which share a similar distribution.
In addition, the word clouds of these abnormality diagnoses are shown in (Fig.~\hyperref[fig:data_stat]{2i}).

\subsection{Developing CT-GLIP Framework}

\noindent \textbf{Problem Formulation}.
The basic motivation for cross-modal contrastive learning is to learn visual concepts from text by training an image encoder $f$ and a text encoder $g$ to maximize the similarity between corresponding image and text feature embeddings while minimizing it for non-corresponding pairs. For a batch of $N$ image-text pairs ($V_n$, $T_n$), we first get the $i$-th normalized visual feature $\bv_i = f(V_i)$ and the $i$-th normalized textual feature $\bt_i = f(T_i)$. Then, the loss for a single pair is shown below:
\begin{align}
    \mathcal{L}_i = -\log \frac{\exp(\bv_i^T \bt_i) / \tau}{\sum^N_{k=1} \exp(\bv_i^T \bt_k) / \tau} -\log \frac{\exp(\bv_i^T \bt_i) / \tau}{\sum^N_{k=1} \exp(\bv_k^T \bt_i) / \tau},
\end{align}
where $\tau$ is a temperature parameter. The total loss is $\mathcal{L} = \frac{1}{N}\sum_{i=1}^N \mathcal{L}_i$.

\noindent \textbf{Model Architecture}.
The CT-GLIP framework is a dual‑path architecture, comprising a 3D vision encoder and a text encoder.
For the 3D vision encoder, we employ both representative CNN- and ViT-based vision encoders, particularly nnUNet~\cite{isensee2021nnu} and MiT~\cite{xie2022unimiss}. 
To keep the low-level semantics, we feed the feature map with the highest resolution into the organ-level average pooling layer. On top of organ-level average pooling, an additional two-layer MLP (hidden layer 768-$d$ with ReLU) is added.
For the text encoder, we adopt ClinicalBERT~\cite{alsentzer2019publicly} as an expert text encoder. We keep the expert text encoder frozen to avoid catastrophic forgetting by CT-specific domain data~\cite{liu2023m}.

\noindent \textbf{Anatomy Contrastive Learning}.
The motivation for anatomy contrastive learning is to learn the basic visual-textual medical concepts, which enable the model to associate the anatomical regions with the corresponding textual descriptions. Following the previous work~\cite{wang2022medclip}, we adopt ClinicalBERT~\cite{alsentzer2019publicly} as an expert text encoder to compute the embedding for text description.
To achieve anatomy contrastive learning, we pair the organ-level visual embeddings and the corresponding grounded textual embeddings. The training procedure is presented in Fig.~\hyperref[fig:method]{3a}.
Specifically, given a CT image $V_i$, the vision encoder projects the CT image into the representation space and produces a feature map $\bv_i$.
Based on the multi-organ segmentation pseudo-label, we apply organ-level average pooling on each segmented organ mask to obtain a set of organ-level features $\{\bz_{ij}\}^M_{j=1}$, where the $M$ refers to the number of organs in the given CT image.
For each organ, we generate its textual description $T_{ij}$ by integrating the specified organ into a predefined template, like ``\texttt{this is a \{organ\} in the CT scan}''.
We then feed the organ descriptions into the expert text encoder to produce organ-level text embedding $\{\bt_{ij}\}^M_{j=1}$.
After that, our training objective $\mathcal{L}_\text{OT}$ is to align organ-text features, as follows:
\begin{equation}
    {\mathcal{L}_\text{anatomy}}_i =  \frac{1}{M} \sum^M_{j=1} \left(-\log \frac{\exp(\bz_{ij}^T \bt_{ij}) / \tau}{\sum^M_{k=1} \exp(\bz_{ij}^T \bt_{ik}) / \tau} -\log \frac{\exp(\bz_{ij}^T \bt_{ij}) / \tau}{\sum^M_{k=1} \exp(\bz_{ik}^T \bt_{ij}) / \tau}\right),
\end{equation}
where the temperature parameter $\tau$ is set as 0.07. Furthermore, to enhance the utilization of the given pseudo-segmentation label $\Tilde{y}$, we introduce an additional segmentation head to predict organs at the pixel level. The segmentation objective $\mathcal{L}_\text{segm}$ is a mixture of cross-entropy loss and dice loss.

\noindent\textbf{Diagnosis Contrastive Learning and Abnormality Dictionary}.
The goal of diagnosis contrastive learning is to integrate the knowledge of abnormality into the 3D VL foundation model.
The training procedure of diagnosis contrastive learning is illustrated in Fig.~\hyperref[fig:method]{3b}.
Similar to anatomy contrastive learning, we first extract organ-level visual features embeddings $\{\bz_{ij}\}^M_{j=1}$ from the given CT image $V_i$.
We then organize $M$ diagnostic descriptions, including $M'$ organ-level real diagnostic descriptions for abnormal organs and $M-M'$ generated descriptions with a predefined template (\eg, ``\texttt{no evident abnormality in \{organ\}}'') for normal organs.
Furthermore, to scale up the diversity and the amount of contrastive pairs~\cite{he2020momentum,oord2018representation} for diagnosis contrastive learning, we introduce an abnormality dictionary storing diverse text descriptions of abnormalities for 104 organs.
In particular, for each normal organ, we look up $T$ abnormal descriptions from the abnormality dictionary and integrate $B = (M-M') \times T$ abnormal descriptions in total.
These $B$ abnormal descriptions provide additional negative pairs for multimodal contrastive learning to distinguish among diseases.
After that, all $M+B$ text descriptions are fed into the expert language model, producing the text embedding $\{t_{ij}\}_{j=1}^{M+B}$.
Given the organ-level paired vision and text embeddings, the training objective of diagnosis contrastive learning is shown below:
\begin{equation}
    {\mathcal{L}_\text{diagnosis}}_i =  \frac{1}{M} \sum^M_{j=1} \left(-\log \frac{\exp(\bz_{ij}^T \bt_{ij}) / \tau}{\sum^{M+B}_{k=1} \exp(\bz_{ij}^T \bt_{ik}) / \tau} -\log \frac{\exp(\bz_{ij}^T \bt_{ij}) / \tau}{\sum^M_{k=1} \exp(\bz_{ik}^T \bt_{ij}) / \tau}\right),
\end{equation}

\noindent\textbf{Overall objective}.
The overall objective of our organ-level vision-language alignment is calculated as the integration of anatomy contrastive loss $\mathcal{L}_\text{anatomy}$, diagnosis contrastive loss $\mathcal{L}_\text{diagnosis}$, and auxiliary cross-entropy loss $\mathcal{L}_{segm}$ (dice loss supervised by pseudo-masks):
\begin{equation}
    \mathcal{L} = \lambda_1\mathcal{L}_\text{anatomy} + \lambda_2\mathcal{L}_\text{diagnosis} + \lambda_3\mathcal{L}_\text{segm},
\end{equation}
where the weights $\lambda_1$,  $\lambda_2$, and $\lambda_3$ are set to 0.5, 0.5, and 1.0, respectively.

\noindent\textbf{Training Details}. For vision–language pretraining, we use a two-stage pipeline. We first pretrain the 3D vision encoder on a segmentation task with pseudo-organ masks generated by TotalSegmentator, providing the model with a basic capability of anatomical discrimination. We then perform grounded contrastive learning for fine-grained VL alignment. Training uses a batch size of 8 and Adam with a weight decay of 3e-5. The learning rate follows cosine decay from 1e-3 to 1e-6. We train for 20 epochs, where the loss has converged. All experiments are conducted on 4$\times$V100 GPUs.

\subsection{Details about Zero-shot Evaluation}
\noindent \textbf{Dataset}.
We evaluate zero-shot performance on a test set of 1,130 samples. For zero-shot organ recognition, the target 104 organs are those segmented by TotalSegmentator~\cite{wasserthal2023totalsegmentator}.
For the zero-shot abnormality detection, we first select the 14 most frequent anatomical regions covering 7 organs, including the spleen, pancreas, aorta, gallbladder, kidney, liver, and lung.
Then, we select the 1-3 most common abnormalities from these 7 organs shown below:
\begin{itemize}
    \item Spleen: splenomegaly and spleen calcification.
    \item Pancreas: acute pancreatitis, chronic pancreatitis, and pancreatic duct stones.
    \item Aorta: arteriosclerosis of the aorta.
    \item Kidney: kidney stone and renal cyst.
    \item Liver: fatty liver, hepatic cyst, and hepatic calcification.
\end{itemize}

\noindent \textbf{Implementation Details}.
Fig.~\hyperref[fig:zeroshot]{4a} illustrates the inference process for zero-shot organ recognition. In particular, we first generate organ descriptions for all 104 organs using a given template. We then convert these descriptions into text embeddings using an expert text encoder. Meanwhile, the corresponding CT scans and multi-organ segmentation are fed into the 3D vision encoder to produce organ-level visual embeddings. The organ label whose text embedding is closest to the grounded visual embedding is then predicted as the most likely category for the organ.
This approach allows CT-GLIP to perform 104-way organ classification tasks on CT scans using just organ descriptions of possible outcomes, enabling accurate and flexible classification without direct training on the task's specific classes.
Fig.~\hyperref[fig:zeroshot]{4c} presents the inference process of zero-shot abnormality detection. For each test CT image, we provide a pair of normal and abnormal text descriptions with corresponding organ segmentation. We assess the similarity between the organ-level grounded visual features and both normal and abnormal textual embeddings for each targeted abnormality. The prediction is made based on the higher similarity score. Clearly, zero-shot abnormality detection operates as a binary classification task.

\subsection{Details about Fine-tuning Evaluation}

\noindent \textbf{Dataset}. For the evaluation of the proposed CT-GLIP in a downstream fine-tuning context, we prepare an in-house dataset encompassing 700 non-contrast CT scans of 700 patients, specifically targeting seven of the most prevalent types of cancer, including lung, breast, liver, esophagus, stomach, colorectum, and pancreas cancer, 100 patients for each type. This dataset is designed to validate the adaptability and performance of our pretrained model in the segmentation and detection of these types of cancer on non-contrast CT scans, which is an emerging and challenging clinical task~\cite{cao2023large}. Seven board-certified radiologists manually annotated the pixel-level mask of the tumors, all confirmed by histopathology. We randomly split the dataset into 448, 112, and 140 cases for training, validation, and test set, respectively.

\noindent \textbf{Implementation Details}.
We employ the same two backbone architectures, \ie, nnUNet~\cite{isensee2021nnu} and the MiT~\cite{xie2022unimiss} network.  For the nnUNet backbone, we use the original training schedule and the self-configured architecture, only with our pretrained model as initialization. The batch size is 8, and we train all experiments for 125K iterations. For the MiT backbone, we add an UNet-style decoder for the segmentation task, fix the MiT encoder for the initial 25K iterations, and tune the whole encoder-decoder network for another 100K iterations. The optimizer for finetuning MiT is RAdam~\cite{liu2019radam} with an initial learning rate of 0.001 and a polynomial learning rate decay.

\section{Declarations}
\subsection{Author Contributions}
Jingyang Lin, Y.X., L.Z., J.Z., K.Y., and L.L. conceived of and designed the study. Y.X., L.Z., and K.C. carried out the data acquisition. Jingyang Lin, Y.X., and L.Z. carried out the data preprocessing. Jingyang Lin and Y.X. developed the AI model. Jingyang Lin, Y.X., and L.Z. analyzed and interpreted the data. Jingyang Lin, Y.X., and L.Z. carried out the statistical analysis. Jingyang Lin and Y.X. wrote the main manuscript. Jingyang Lin prepared figures 1-5. Jingyang Lin, Y.X., L.Z., L.L., and Jiebo Luo reviewed and revised the manuscript.
\subsection{Competing Interest}
The authors declare no competing interests.
\subsection{Acknowledgements}
We thank the Department of Radiology, Shanghai Institute of Pancreatic Disease, for collecting the paired CT–report dataset and conducting careful data preprocessing and de-identification. We also thank DAMO Academy, Alibaba Group, for providing GPU resources.
\subsection{Funding Statement}
Funding: Not applicable.
\subsection{Data Availability}
The datasets analyzed in this study are not yet publicly available due to restrictions imposed by the respective IRBs, which are responsible for ensuring the protection of patient privacy and compliance with ethical standards. Access to the original data is, therefore, limited and cannot be shared openly. 
Access is conditioned on the IRB approvals and execution of a data-use agreement. Once the access is granted, the dataset will be available for noncommercial academic purposes only.


\bibliography{sn-bibliography}

\newpage

\appendix

\section{Supplementary Materials}\label{sec:appendix}

\subsection{Evaluation Metrics}

\noindent \textbf{Zero-shot Organ Recognition}.
Zero-shot organ recognition is a 104-way classification task. We evaluate this task using Top-1 accuracy, which is calculated by the proportion of test cases where the highest-scoring prediction $\hat{y}_i$ matches the ground-truth label $y_i$:
\begin{equation}
    \text{Top-1 Acc} \;=\; \frac{1}{N}\sum_{i=1}^{N} \mathbf{1}\!\left[ \hat{y}_i = y_i \right].
\end{equation}

\noindent \textbf{Zero-shot Abnormality Detection}.
Zero-shot abnormality detection (binary classification task) is assessed by Positive Predictive Value (PPV), F1 score, and AUC:
\begin{align}
\text{PPV} &= \frac{\mathrm{TP}}{\mathrm{TP}+\mathrm{FP}}, \\
\text{Sensitivity} &= \frac{\mathrm{TP}}{\mathrm{TP}+\mathrm{FN}}, \\
\text{F1} &= \frac{2\,\text{PPV}\cdot\text{Sensitivity}}{\text{PPV}+\text{Sensitivity}}
           = \frac{2\,\mathrm{TP}}{2\,\mathrm{TP}+\mathrm{FP}+\mathrm{FN}}.
\end{align}
We compute AUC by ranking predictions by decreasing abnormality score, varying the decision threshold to obtain TPR (true positive rate) and FPR (false positive rate) across all points to trace the ROC curve, and estimating the enclosed area.

\noindent \textbf{Tumor Detection}.
As for the patient-level detection of each type of tumor (the presence or absence of each tumor), we use the 3D volume of the respective tumor as the score for the computation of AUC~\cite{zhu2019multi}.

\noindent \textbf{Tumor Segmentation}.
The performance is evaluated by the Dice-Sørensen
Coefficient (DSC) for tumor segmentation:
\begin{equation}
    \mathrm{DSC}(\hat{\mathcal{Y}},\mathcal{Y}) \;=\; \frac{2\,|\hat{\mathcal{Y}} \cap \mathcal{Y}|}{|\hat{\mathcal{Y}}| + |\mathcal{Y}|},
\end{equation}
where $\hat{\mathcal{Y}}$ is the set of foreground voxels predicted by the model, and $\mathcal{Y}$ is the set of foreground voxels in the ground-truth annotation.

\subsection{Raw Results}
For completeness, raw evaluations for all models in Figs.\hyperref[fig:zeroshot]{4} and \hyperref[fig:ft]{5} are provided in Tables\ref{tab:zeroshot} and \ref{tab:finetune}, respectively.

\begin{table}[t]
\centering
\caption{The performance of zero-shot organ recognition and abnormality detection.  ACL indicates anatomy contrastive learning, DCL refers to diagnosis contrastive learning, and A-Dict denotes abnormality dictionary. Top-1 accuracy, PPV (Positive Predictive Value), Sensitivity, F1 score, and AUC are shown in \%.}
\label{tab:zeroshot}
\begin{tabular}{llcccccc}
\toprule[1.2pt]
\multirow{3}{*}{Method} & & \multicolumn{1}{c}{\textbf{Zero-shot}} & \multicolumn{4}{c}{\textbf{Zero-shot}}    \\ 
&  & \multicolumn{1}{c}{\textbf{Organ Recognition}} & \multicolumn{4}{c}{\textbf{Abnormality Detection}}    \\ 
 \cmidrule(r){3-3} \cmidrule(r){4-7}
                      &   & Top-1 Acc $\uparrow$    & PPV $\uparrow$   & Sensitivity $\uparrow$ & F1 $\uparrow$    & AUC $\uparrow$  \\ \midrule[1.2pt]
\multicolumn{7}{>{\columncolor[gray]{.85}}l}{\textbf{CNN-based architecture}: nnUNet} \\
Vanilla CLIP~\cite{radford2021learning} &               &    0.01  &     32.75    &      35.19  &   33.93  &     52.23     \\ 
\midrule
\multirow{3}{*}{CT-GLIP} &+ACL                 &   0.03   &   35.24     &   70.66   &  47.02     &   66.00  \\    
  &+ACL +DCL                &   86.92   &    39.07     &    64.11     &   48.60   &  66.76  \\    
 &+ACL +DCL +A-Dict                 &   86.24   &    39.24     &    72.85     &   49.02   &  68.63  \\    
\midrule[1.2pt]
\multicolumn{7}{>{\columncolor[gray]{.85}}l}{\textbf{ViT-based architecture}: MiT} \\
Vanilla CLIP~\cite{radford2021learning}   &               &  0.01   &      34.01   &    40.43      &    36.94   &       52.37      \\ 
\midrule
\multirow{3}{*}{CT-GLIP} & +ACL         &   0.07   &    37.65     &   74.24      &  49.96    &  69.27  \\      
& +ACL +DCL         &   85.46   &   38.24     &  77.43       &   51.19  &  70.12  \\    
& +ACL +DCL  +A-Dict  & 84.93  &   39.47   &   78.59  &  52.55    &  71.90  \\    
\bottomrule[1.2pt]
\end{tabular}
\end{table}

\begin{table}[t]
\centering
\caption{The performance of downstream fine-tuning on the task of cancer screening of pancreas (Pan), breast (Bre), stomach (Sto), colorectum (Col), lung, esophagus (Eso), and liver (liv). Tumor segmentation is evaluated via DSC (\%), and the performance of cancer screening is evaluated via AUC (\%).}
\label{tab:finetune}
\begin{tabular}{ll@{\hspace{0.7cm}}c@{\hspace{0.7cm}}c@{\hspace{0.7cm}}c@{\hspace{0.7cm}}c@{\hspace{0.7cm}}c@{\hspace{0.7cm}}c@{\hspace{0.7cm}}c@{\hspace{0.7cm}}c}
\toprule[1.2pt]
Metric & Method & Pan & Bre & Sto & Col & Lung & Eso & Liv & Mean \\ 
\midrule[1.2pt]
\multirow{8}{*}{DSC $\uparrow$} & \multicolumn{9}{>{\columncolor[gray]{.85}}l}{\textbf{CNN-based architecture}: nnUNet} \\
& Scratch & 38.77 & 13.20 & 20.36 & 24.18 & 40.45 & 52.94 & 19.25 & 29.88 \\
& Vanilla CLIP~\cite{radford2021learning} & 50.58 & 19.28 & 19.98 & 24.31 & 47.15 & 56.45 & 16.43 & 33.45 \\
& CT-GLIP (ours)    & 52.42 & 20.59 & 23.13 & 26.16 & 48.89 & 53.25 & 18.44 & 34.70 \\ 
\cmidrule(r){2-10} & \multicolumn{9}{>{\columncolor[gray]{.85}}l}{\textbf{ViT-based architecture}: MiT} \\
& Scratch & 35.18 & 19.13 & 0.00  & 18.56 & 11.12 & 40.32 & 34.41 & 22.68 \\
& Vanilla CLIP~\cite{radford2021learning} & 36.02 & 19.96 & 17.47 & 28.78 & 33.39 & 50.19 & 28.75 & 30.65 \\
& CT-GLIP (ours) & 39.85 & 22.84 & 27.23 & 34.15 & 42.32 & 46.60 & 37.39 & 35.77 \\
\midrule[1.2pt]
\multirow{8}{*}{AUC $\uparrow$} & \multicolumn{9}{>{\columncolor[gray]{.85}}l}{\textbf{CNN-based architecture}: nnUNet} \\
 & Scratch & 92.19 & 70.23 & 83.48 & 90.72 & 74.91 & 100.00 & 63.31 & 82.12 \\
& Vanilla CLIP~\cite{radford2021learning}    & 96.69 & 80.97 & 83.17 & 86.99 & 92.05 & 100.00 & 71.39 & 87.32 \\
&   CT-GLIP (ours)    & 97.73 & 81.49 & 90.64 & 90.54 & 92.43 & 93.70 & 80.35 & 89.55 \\
\cmidrule(r){2-10} & \multicolumn{9}{>{\columncolor[gray]{.85}}l}{\textbf{ViT-based architecture}: MiT} \\

& Scratch & 97.63 & 77.91 & 50.00 & 90.26 & 74.05 & 98.74 & 77.99 & 80.94 \\
&     Vanilla CLIP~\cite{radford2021learning}    & 90.10 & 85.83 & 78.40 & 95.65 & 78.45 & 99.55 & 81.03 & 87.00 \\
&  CT-GLIP (ours)    & 91.48 & 81.31 & 87.79 & 95.03 & 85.76 & 96.35 & 82.46 & 88.60 \\
\bottomrule[1.2pt]
\end{tabular}
\end{table}

\end{document}